\documentclass[letterpaper, 10 pt, conference]{ieeeconf}  

\makeatletter
\let\NAT@parse\undefined
\makeatother
\usepackage[colorlinks,urlcolor=blue,linkcolor=blue,citecolor=green]{hyperref}
\usepackage{bookmark}
\usepackage{color,array}
\usepackage{cite}
\usepackage{graphicx}
\usepackage{upgreek}
\usepackage{booktabs}  
\usepackage{threeparttable}  
\usepackage{multirow}
\usepackage{epstopdf}
\usepackage{amssymb}
\usepackage{algorithm}
\usepackage{algorithmicx}
\usepackage{algpseudocode}
\usepackage{setspace}
\usepackage{url}
\usepackage{subfig}
\usepackage{gensymb}
\usepackage{soul}
\usepackage{threeparttable}
\soulregister\cite7 
\soulregister\citep7 
\soulregister\citet7 

\IEEEoverridecommandlockouts                              

\title{\LARGE \bf
On Deep Recurrent Reinforcement Learning for Active Visual Tracking of Space Noncooperative Objects
}

\author{Dong Zhou$^{1}$, Guanghui Sun$^{1, *}$, Zhao Zhang$^{1}$ and Ligang Wu$^{1}$
\thanks{This work was kindly supported by the National Key R\&D Program of China through grant 2019YFB1312001.}
\thanks{$^{1}$D. Zhou, G. Sun, Z. Zhao, and L. Wu are with the Department of Control Science and Engineering, Harbin Institute of Technology, Harbin, China, 150001.}
\thanks{$^{*}$Correspoding author: G. Sun({\tt\small guanghuisun@hit.edu.cn})}%
}

\begin{document}

\maketitle
\thispagestyle{empty}
\pagestyle{empty}

\newcommand{\tabincell}[2]{\begin{tabular}{@{}#1@{}}#2\end{tabular}}

\begin{abstract}
   Active tracking of space noncooperative object that merely relies on vision camera is greatly significant for autonomous rendezvous and debris removal. Considering its Partial Observable Markov Decision Process (POMDP) property, this paper proposes a novel tracker based on deep recurrent reinforcement learning, named as RAMAVT which drives the chasing spacecraft to follow arbitrary space noncooperative object with high-frequency and near-optimal velocity control commands. To further improve the active tracking performance, we introduce Multi-Head Attention (MHA) module and Squeeze-and-Excitation (SE) layer into RAMAVT, which remarkably improve the representative ability of neural network with almost no extra computational cost. Extensive experiments and ablation study implemented on SNCOAT benchmark show the effectiveness and robustness of our method compared with other state-of-the-art algorithm. The source codes are available on \url{https://github.com/Dongzhou-1996/RAMAVT}.
\end{abstract}


\section{INTRODUCTION\label{section1}}
With the rapid development of aerospace technology, space noncooperative object active visual tracking that drives the chasing spacecraft or space manipulator to pursue any specific noncooperative target by merely using vision camera has attracted extensive attentions. It is essential to intelligent on-orbit service such as autonomous rendezvous \cite{kirkhovellDeepReinforcementLearning2021,zhaoImagebasedControlRendezvous2021,zhou2022space}, space debris removal\cite{huangDexterousTetheredSpace2017, forshaw2020active, aglietti2020active}, and malfunctioning satellite maintenance \cite{flores-abadReviewSpaceRobotics2014, fourie2014flight,li2019orbit}.


Benefitting from the powerful deep reinforcement learning (DRL) that has achieved great successes in many fields like video game \cite{mnihHumanlevelControlDeep2015}, Go\cite{silverMasteringGameGo2016}, autonomous driving \cite{kiran2021deep}, and robotic manipulation \cite{singh2021reinforcement}, more and more active visual trackers \cite{luoEndtoendActiveObject2018, cruciataUseDeepReinforcement2021, devoEnhancingContinuousControl2021, zhongADVATAsymmetricDueling2021, heejinjeongDeepReinforcementLearning2021, zhong2021towards, tiritirisTemporalDifferenceRewards2021a, dionigiEVATAsymmetricEndtoEnd2022, xiAntiDistractorActiveObject2022} have been proposed in an end-to-end manner, which can learn global optimal policy after training with millions of trial-and-errors experiences.

\begin{figure}[t]
   \centering
   \includegraphics[width= 0.32\textwidth]{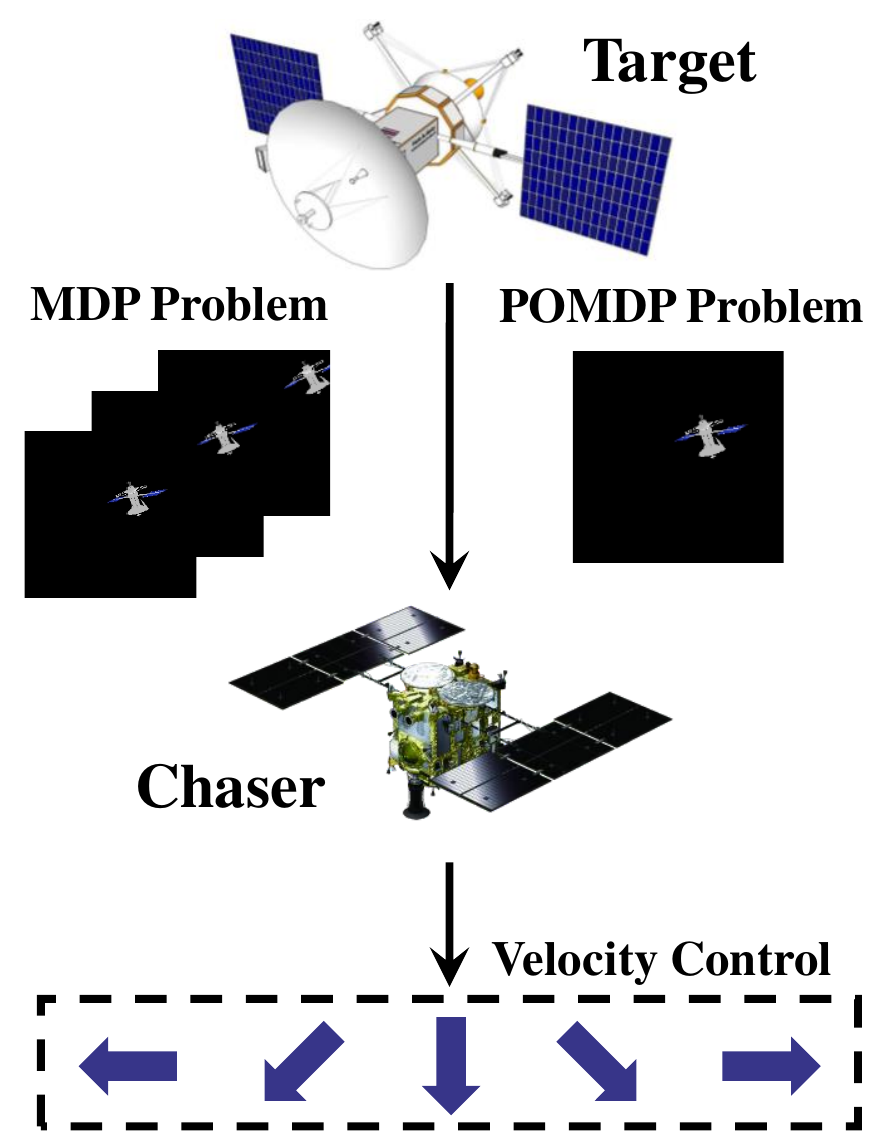}
   \caption{Space noncooperative object active visual tracking}
   \label{fig1}
\end{figure}

In our preliminary work \cite{zhou2022space}, we presented the SNCOAT benchmark \cite{sncoat} and the first active visual tracker, DRLAVT in aerospace domain. It achieves impressive tracking performance in velocity control mode by stacking multiple frames as an input. However, the stacking mechanism not only decreases the control bandwidth severely, but also makes active tracker more vulnerable to perturbations (e.g., image blur, actuator noise, computational delay). 

To this end, we take the POMDP property of space noncooperative object tracking into consideration. A novel active visual tracker based on deep recurrent reinforcement learning, RAMAVT is proposed in this paper, which can drive the chasing spacecraft to pursue arbitrary target with high-frequency and near-optimal velocity control commands. Our method features the accurate perception of target position and velocity, even though taken one image as input per time, which benefits from the recurrent neural network (RNN) in RAMAVT architecture that establishes the relationship between long-term temporal sequence.

Intuitively, it is totally enough for active visual tracker that focus attention on partial regions or partial channels of feature tensor to achieve the information of space noncooperative target. To this end, we adopt the multi-head attention module \cite{vaswani2017attention} and Squeeze-and-Excitation layer \cite{hu2020squeeze} in RAMAVT, which only increase small number of model parameters but significantly improve the representative ability of neural network. In addition, some data augmentation methods \cite{laskin2020reinforcement} are also involved to enhance the efficiency of learning process and generalization ability of active tracker.

\begin{figure*}[t]
   \centering
   \includegraphics[width= 0.9\textwidth]{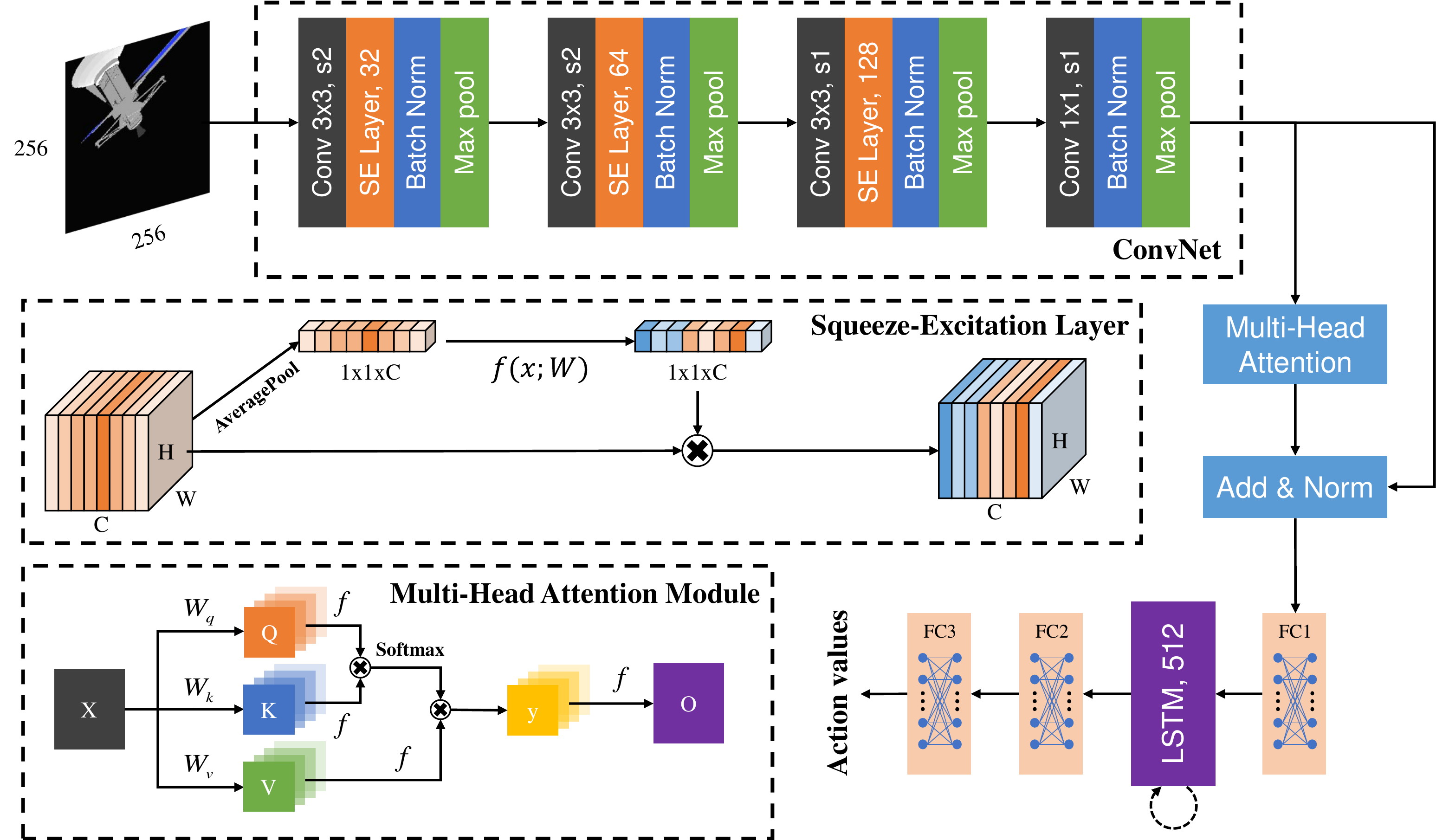}
   \caption{The architecture of RAMAVT, which directly maps an image to optimal velocity control command attributed to the LSTM module that establish long-term relationship in temporal sequence. RAMAVT also adopts MHA module \cite{vaswani2017attention} and SE layer \cite{hu2020squeeze} to improve the representative ability of neural network.}
   \label{fig2}
\end{figure*}

The contributions of our work in this paper are summarized as following: 
\begin{itemize}
   \item We propose a novel active visual tracker based on deep recurrent reinforcement learning, RAMAVT which achieves excellent performance compared to other state-of-the-art methods.
   \item Multi-head attention module and SE layer are adopted into RAMAVT, combined with data augmentation, of which effectiveness has been proved by detailed ablation study. 
\end{itemize}

This work proceeds as follows: Section \ref{section2} introduces some related works about space noncooperative object active visual tracking. Section \ref{section3} describes our RAMAVT method in detail. The experiments and analysis are given in Section \ref{section4}. Finally, we make a conclusion in Section \ref{section5}.


\section{Related Work\label{section2}}
Visual object tracking is a hot topic in computer vision society, which has wide applications in civil, military, and aerospace fields. In recent years, many efforts have been devoted to studying passive methods \cite{chen2021transformer, ondrasovicSiameseVisualObject2021,zhou2DVisionbasedTracking2021, hu2022global, dunnhofer2022visual}, which assume that the target is always within the field of view (FOV) of vision camera. This severely limits the possibility to apply visual object tracking methods in many real-world scenarios, especially for aerospace applications where the target often maneuvers in 6-Degrees-of-Freedom (DoF) and the low-resolution camera mounted on spacecraft only has small FOV.

Therefore, active visual tracking \cite{dongAutonomousRoboticCapture2016, sunAdaptiveRelativePose2018, liuRobustAdaptiveRelative2020, luoEndtoendActiveObject2018, cruciataUseDeepReinforcement2021, devoEnhancingContinuousControl2021, zhongADVATAsymmetricDueling2021, heejinjeongDeepReinforcementLearning2021, zhong2021towards, tiritirisTemporalDifferenceRewards2021a, dionigiEVATAsymmetricEndtoEnd2022, xiAntiDistractorActiveObject2022} has achieved more and more concerns, which not only identifies the target but also changes the pose of the chaser in real-time to keep view contact with the target. Traditional active visual trackers often adopt PBVS or IBVS framework of which modules (e.g. key-points detection, feature matching, pose estimation, and controller design) are optimized separately. 

In the paper \cite{dongAutonomousRoboticCapture2016}, a PBVS algorithm that guide space robotic manipulator to grasp noncooperative target was proposed, in which photogrammetry and adaptive extended Kalman filter are used to predict 6-DoF pose of target. However, this work is unadaptable to complex space environment as the same as other traditional active visual trackers. We also proposed a novel PBVS tracker in our preliminary work \cite{zhou2022space} that adapts state-of-the-art 2D monocular tracking method, SiamRPN \cite{liHighPerformanceVisual2018}, which achieved fairly good active tracking performance on SNCOAT benchmark but make concession in real-time capability.

Deep reinforcement learning that learns optimal action policy in an end-to-end manner with millions of trial-and-errors experiences has made great contributions in many fields, such as video game\cite{mnihHumanlevelControlDeep2015}, Go\cite{silverMasteringGameGo2016}, autonomous driving\cite{kiran2021deep}, and robotic manipulation \cite{singh2021reinforcement}, which provides a novel perspective for active visual tracking. In recent years, many DRL-based active visual trackers were proposed \cite{luoEndtoendActiveObject2018, cruciataUseDeepReinforcement2021, devoEnhancingContinuousControl2021, zhongADVATAsymmetricDueling2021, heejinjeongDeepReinforcementLearning2021, zhong2021towards, tiritirisTemporalDifferenceRewards2021a, dionigiEVATAsymmetricEndtoEnd2022, xiAntiDistractorActiveObject2022}, most of which aim at terrestrial targets and only deploy on unmanned ground vehicle (UGV). 

Luo et al. \cite{luoEndtoendActiveObject2018} proposed the first end-to-end active visual tracker based on A3C \cite{mnihAsynchronousMethodsDeep2016} that can only pursue two person models walking along fixed trajectory in two types of environments. In addition, The training progress takes up several days to achieve nice tracking performance, even with very low-resolution image (84$\times$84$\times$3). The paper \cite{tiritirisTemporalDifferenceRewards2021} presented a temporal difference-based reward function adopted in PPO learning framework \cite{schulmanProximalPolicyOptimization2017}, which effectively decreases the distance error between the chasing and target UGVs. However, this agent was only trained with single target in simulation environment. It was almost overfitted and impossible to track another target in real-world scenario. In contrast, our method is trained with 12 types of space noncooperative objects including space stations, satellites, asteroids, rockets, and return capsules, which successfully guarantees the generalization ability of RAMAVT.

Those algorithms mentioned above assume that the initial position of target is within the active tracker's FOV, which is a demanding condition for real application. To this end, Jeong et al. \cite{heejinjeongDeepReinforcementLearning2021} extended the active visual tracking problem involved navigation, exploration and in-sight tracking and proposed the active tracking target network (ATTN) that learns a unified policy to track agile and anomalous object with partially known target model. This method features the incorporation that feeds egocentric maps and visit frequency to the convolutional neural network (CNN), which formulates the active visual tracking task as Markov Decision Process (MDP). In \cite{dionigiEVATAsymmetricEndtoEnd2022}, Dionigi et al. also presented the DRL-based E-VAT model consisted of target-detection network and exploration-and-tracking network, which can explore the environment and track the target autonomously.

Compared with terrestrial targets, the active visual tracking tasks of space noncooperative objects are more challenging: 1) the tracked target often maneuvers agilely with complex 6-DoF trajectory; 2) less prior knowledge are available, such as geometry, texture, kinematic and dynamic parameters; 3) the images captured by vision camera on spacecraft are often low-quality, because of low resolution, small FOV, camera motion, and illumination variance. 

In our preliminary work \cite{zhou2022space}, we propose the first active tracker DRLAVT in aerospace domain, of which performance has a large room to be improved. The partially observable problem for active visual tracking was avoided by frame-stack mechanism, however, it severely decreases control bandwidth and tracking performance. To this end, we propose a novel active visual tracker based on deep recurrent reinforcement learning that directly maps one image to optimal velocity control command, benefitted from long-term temporal relationship established by RNN. In addition, the MHA module and SE layer are introduced to further improve network representative ability. 

\begin{table*}[t]
   \centering
   \caption{Active Tracking Performance Comparison}
   \label{table1}
   \begin{threeparttable}
   \begin{tabular}{ccccccccccc}
      \toprule
      \multirow{2}{*}{Name} & \multicolumn{3}{c}{Input Format} & \multicolumn{3}{c}{Episode Length} & \multicolumn{3}{c}{Episode Reward} & Speed \\
      \cline{2-10}
      & RGBD & Depth & Color & Avg & Min & Max & Avg & Min & Max & (Hz) \\
      \midrule
      Random & - & - & - & 152.2 & 21 & 385 & -1545.2 & -3214.8 & -178.4 & 42703.5 \\
      \midrule
      \multirow{3}{*}{DRLAVT} & $\surd$ & - & - & 857.9 & 6 & 1000 & -268.9 & -2341.4 & 386.4 & 63.1 \\
       & - & $\surd$ & - & 901.1 & 6 & 1000 & \textcolor{green}{430.3} & \textcolor{green}{-55.69} & 382.2 & 66.6 \\
       & - & - & $\surd$ & 841.6 & 15 & 1000 & -201.9 & -1798.7 & 320.4 & 68.9 \\
      \midrule
       \multirow{3}{*}{RAMAVT} & $\surd$ & - & - & 952.4 & 41 & 1000 & -398.5 & -1523.2 & \textcolor{green}{481.4} & 202.7  \\
      & - & $\surd$ & - & \textcolor{green}{959.1} & \textcolor{green}{162} & 1000 & -59.4 & -800.5 & 445.4 & \textcolor{green}{216.1} \\
      & - & - & $\surd$ & 685.3 & 43 & 1000 & -1755.1 & -4097.5 & 434.7 & 210.9 \\
      \bottomrule
   \end{tabular}

   \begin{tablenotes}
   \footnotesize
   \item \rightline{(The best scores are highlighted in \textcolor{green}{green})}
    \end{tablenotes}
   \end{threeparttable}
\end{table*}

\section{Proposed Method\label{section3}}
In this section, we formulate the active visual tracking problem of space noncooperative object and describe our RAMAVT algorithm thoroughly.  

\subsection{Problem Formulation}
The task of space noncooperative object active visual tracking involves the chaser mounted with vision camera and the moving target with no prior information, where the previous one should change its pose by using images to reduce the error $e_t$ which is formulated as follows:
\begin{equation}
   e_t = \left\| r^{B}_{T}(t) - r^{\ast} \right\|_2 \label{eq1}
\end{equation}
in which  $r^{B}_{T}(t)$ is the 3-D position of the target in the body-frame of the chaser at \emph{t}th timestep, and $r^{\ast}$ denotes the expected distance between the chaser and target. In this work, $r^{\ast}$ is set to $\left\{0,0,5\right\}$.

To complete this task with DRL-based method, it can be further described as a POMDP problem. At \emph{t}th timestep, the state of the target $s_t \in \mathcal{S}$ is observed as $o_t \in \mathcal{O}$ by agent with vision camera. Then, the agent takes action $a_t \in \mathcal{A}$ following a policy, such as the greedy policy $a_t = \mathop{\max }\limits_{a \in \mathcal{A}} Q(o_t, a)$ that is adopted in this article. After that, the agent would receive a reward $r_t$ from the environment which is generated by a reward function. The definition of our reward function is inherited from the paper \cite{zhou2022space}, which includes a visible term $r_{vis}$ and a distance penalty term $r_{dist}$. The partially observable problem means $o_t \neq s_t$, that is, the agent can not accurately perceive actual state of the target, especially for the velocity.

\subsection{RAMAVT Algorithm}
The POMDP problem mentioned above makes an end-to-end active visual tracker difficult to approximate optimal action value function $Q^{\ast}(o_t, a_t)$. To this end, we propose a new deep Q-network architecture based on DRQN \cite{hausknecht2015deep}, as shown in Fig. \ref{fig2}, which can establish the long-term relationship between temporal sequence and directly map one image to optimal velocity control command. 

Meanwhile, some additional SE layers \cite{hu2020squeeze} and Multi-head attention module \cite{vaswani2017attention} are also introduced to improve the representative ability of deep Q-network and approximate better action value function $Q(o_t, a_t)$. The SE layer features the modelling of the channel-wise interdependencies of feature tensors with low computational cost, of which schematic is illustrated at the middle part of Fig. \ref{fig2}. In this work, We place SE layer behind every convolutional layer in ConvNet backbone.

In recent years, the self-attention mechanism derived from natural language processing filed has been widely applied to computer vision, which significantly increases the representation of neural network to images. The MHA module, shown in the bottom of Fig. \ref{fig2}, is one of the most famous self-attention method representing the interrelationship of different positions in one image. The basic of MHA is the scaled dot-product attention algorithm:
\begin{equation}
   y_i = \textbf{softmax}\left( \frac{Q_i K^{T}_i}{\sqrt{d_k}}\right)V_i
\end{equation}
where $Q_i =W_q \cdot x_i$, $K_i = W_k \cdot x_i$, and $V_i = W_v \cdot x_i$ are the three feature vectors computed by different full-connected (FC) layers fed with the same input $x_i$, and $d_k$ denotes the dimension of feature vector $K_i$. Based on this, the MHA can be formulated as:
\begin{equation}
   O = W_o \cdot \textbf{Concat}\left\{ y_1, y_2, \cdots, y_N \right\}
\end{equation}
in which, $N$ is the number of heads that attend to information from different representation subspaces at different positions in parallel. We adopt $N=8$ heads in this work.

Finally, we train the RAMAVT model with loss function $\mathcal{L}(\theta)$ defined as follows:  
\begin{equation}
   \mathcal{L}(\theta) = \mathbb{E}_{(o, a, r, o') \sim U(H)}\left[(y - Q(o, a; \theta))^2\right]
\end{equation}
where the training data $(o, a, r, o')$ is uniformly sampled from the hierarchical memory pool $H$ proposed by us that is more suitable for deep recurrent reinforcement learning methods. $y = r + \gamma  \mathop{\max }\limits_{a' \in \mathcal{A}} Q(o', a'; \theta^{-})$ is the Temporal-difference (TD) target estimated by the target network $\theta^{-}$.

\begin{table}[t]
	\centering
	\caption{Training Configurations}
	\label{table4}
	\begin{tabular}{lcl}
		\toprule
		Params & Value & Note \\
		\midrule
		replay buffer & 50000 & \tabincell{l}{The size of replay pool }\\
		initial buffer & 10000 & \tabincell{l}{The number of initial \\ experiences} \\
		episode num & 300 & \tabincell{l}{The number of episodes \\ used to train Q-network}\\
		max episode len & 1000 & \tabincell{l}{The max length of one \\ episode, but if target is \\ lost, episode will be over}\\
		update interval & 10 & \tabincell{l}{The update interval of \\ target network }\\
		gamma & 0.99 & \tabincell{l}{Rewards discount factor}\\
		\bottomrule
	\end{tabular}

\end{table}

\section{Experiment\label{section4}}
In this section, we first validate the active tracking performance of RAMAVT by using the evaluation toolkit provided in SNCOAT benchmark \cite{sncoat}. Then, sufficient ablation studies on RAMAVT are also implemented to show the effectiveness of our method. Finally, we further explore the working mechanism of active visual trackers. All the trackers follows the same training configurations listed in Table \ref{table4}. The experimental platform is HPC server equipped with Intel Xeon@E5 2650v4 CPU and Nvidia Tesla P100 GPU.

\begin{figure}[t]
   \centering
   \includegraphics[width=0.45\textwidth]{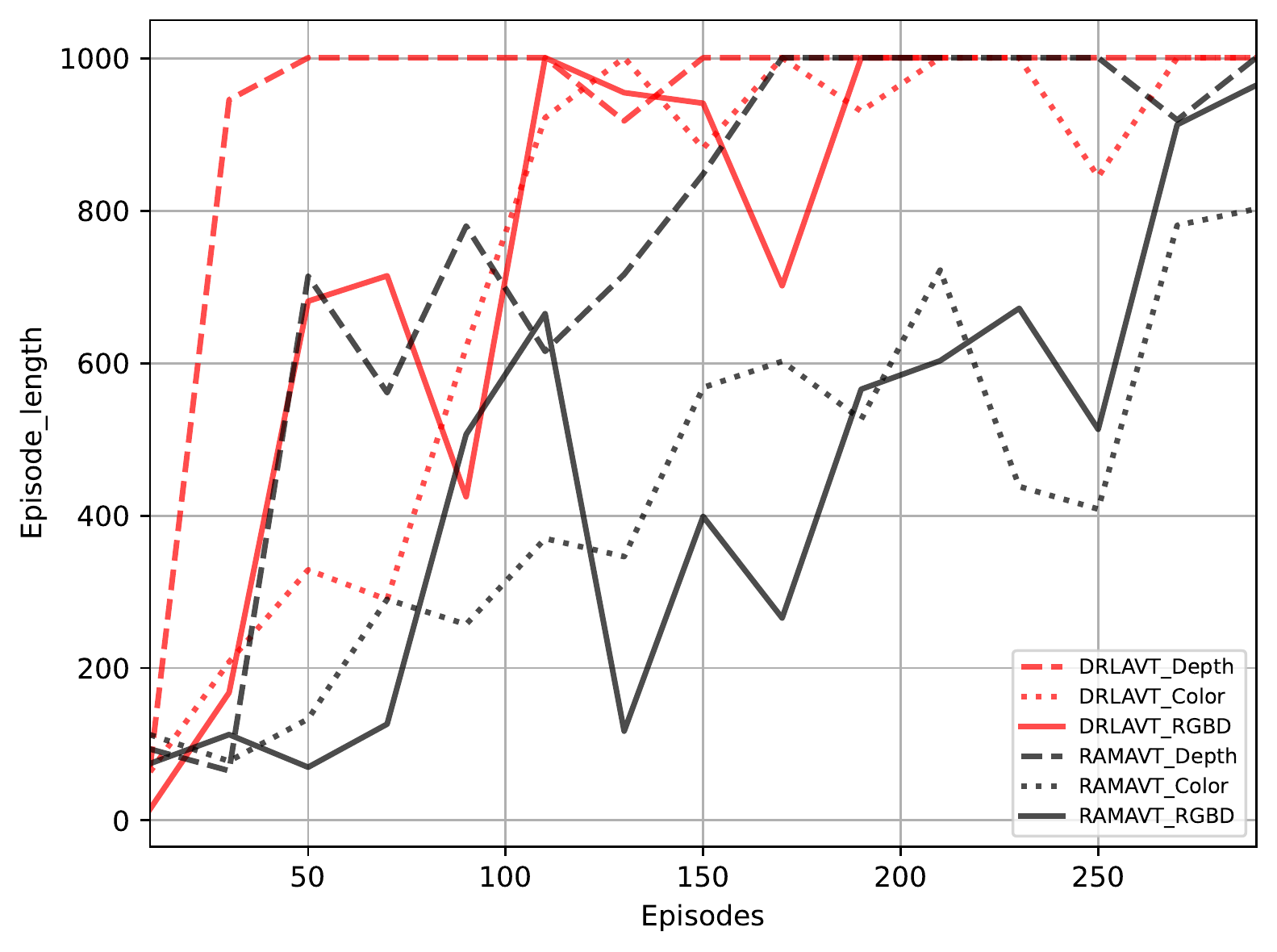}
   \caption{The training curves of different active visual trackers}
   \label{fig3}
\end{figure}

\begin{figure*}[t]
   \centering
   \begin{tabular}{p{0.42\textwidth}<{\centering} p{0.3\textheight}<{\centering}}
      \subfloat[active tracking trajectories]{
      \includegraphics[width = 0.20\textwidth, height=0.15\textheight]{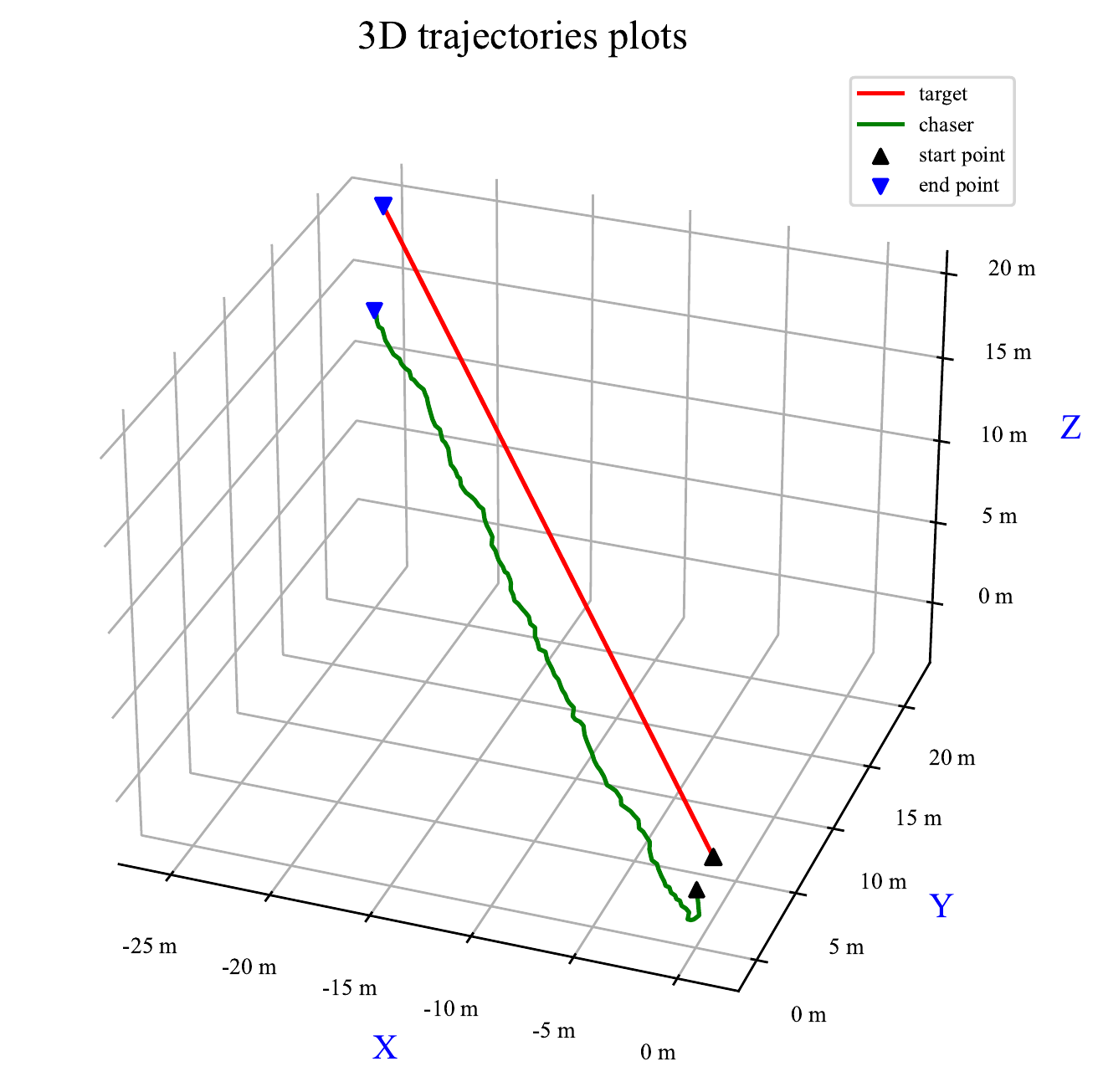}
      \includegraphics[width = 0.20\textwidth, height=0.15\textheight]{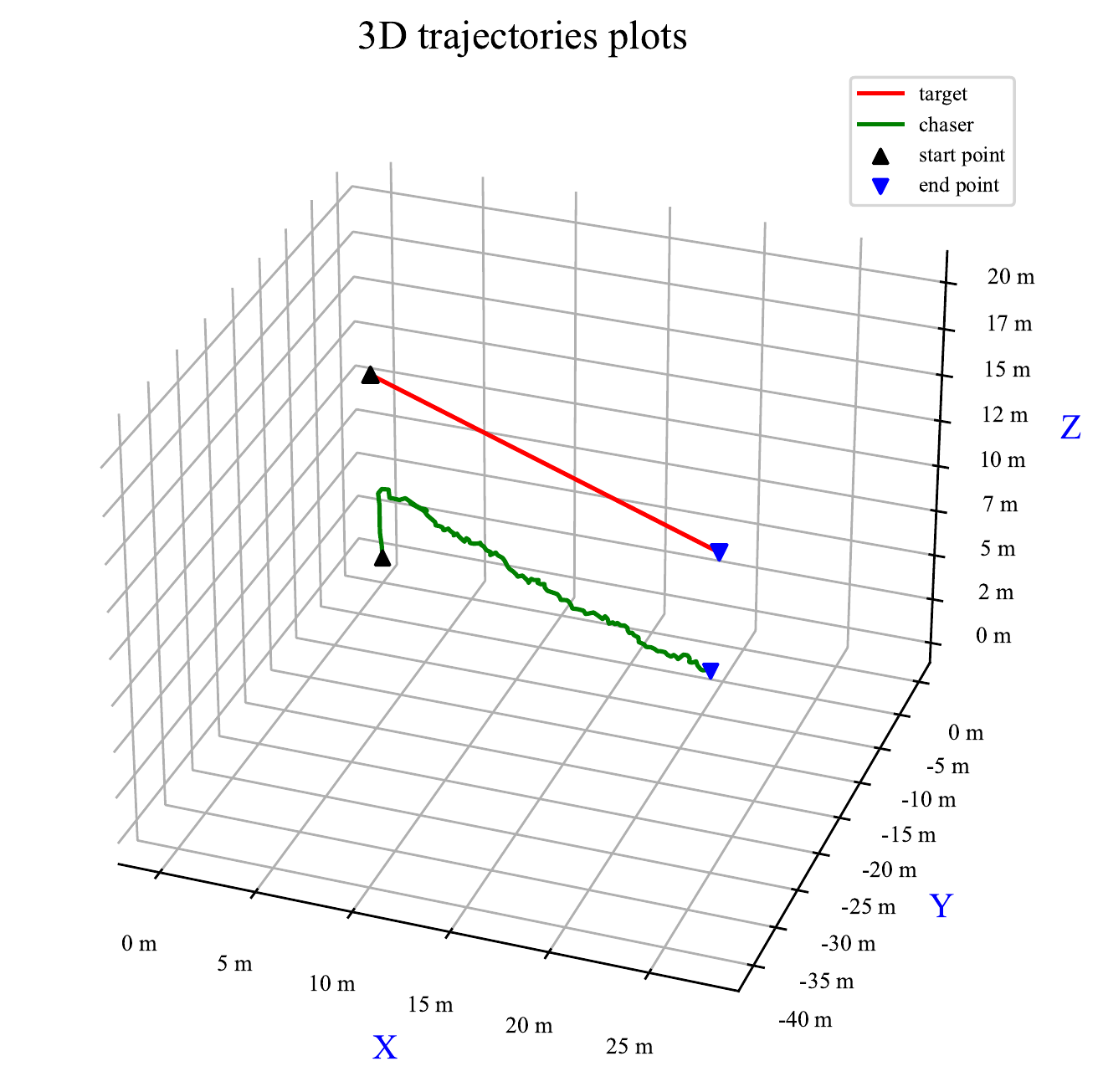}
   \label{fig5_a}
   } & 
   \multirow{2}{*}{
      \subfloat[active tracking errors in one trajectory]{
         \includegraphics[width = 0.45\textwidth, height=0.26\textheight]{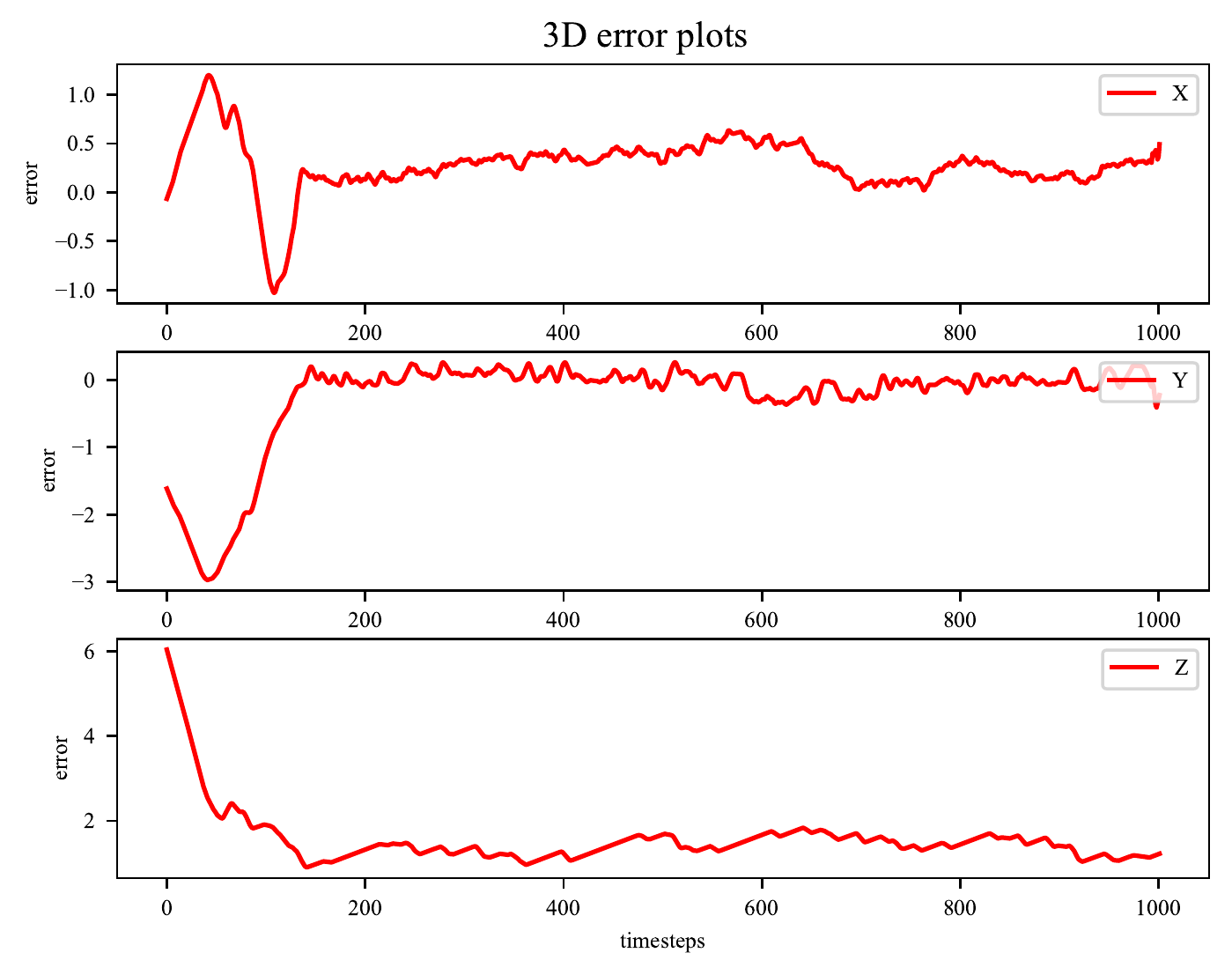} 
      \label{fig5_b}}
   }
   \\
   \subfloat[observed depth maps in one trajectory]{
      \includegraphics[width = 0.45\textwidth]{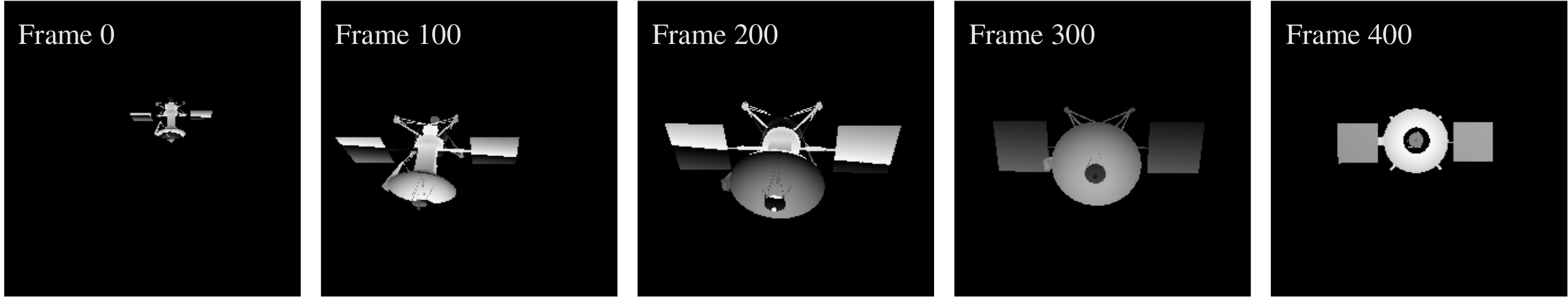} \label{fig5_c}
   } &
   \\
   \end{tabular}
   \caption{The results of RAMAVT with Depth image.}
   \label{fig5}
\end{figure*}

\begin{table}
   \centering
   \caption{The Robustness Evaluation Under Different Perturbations}
   \label{table2}
   \begin{threeparttable}
   \begin{tabular}{cccccc}
      \toprule
      \multirow{2}{*}{Name} & \multicolumn{3}{c}{Perturbations} & \multicolumn{2}{c}{Metrics} \\ 
      \cline{2-6}
      & \tabincell{c}{Actuator\\Noise}  & \tabincell{c}{Time\\Delay}  &  \tabincell{c}{Image\\Blur}   & AEL & AER \\
      \midrule
      DRLAVT & $\surd$ & $\surd$ & $\surd$ & \textcolor{red}{456.3} & -758.6 \\
      \midrule
      \multirow{5}{*}{RAMAVT}& $\surd$ & - & - & 876.5 & -438.9 \\
      & - & $\surd$ & - & 596.6 & -793.3 \\
      & - & - & $\surd$ & 793.3 & -810.1 \\
      & $\surd$ & $\surd$ & $\surd$ & 580.8 & \textcolor{red}{-971.8} \\
      \cline{2-6}
      & - & - & - & 952.4 & -398.5 \\

      \bottomrule
   \end{tabular}

   \begin{tablenotes}
      \footnotesize
      \item \rightline{(The worst scores are highlighted in \textcolor{red}{red})}
   \end{tablenotes}
   \end{threeparttable}
\end{table}

\subsection{RAMAVT Performance}
We train the agent with 12 types of space noncooperative objects and evaluate it on other 6 different targets, including asteroids, satellite, rockets, space station, and return capsule. Some data augmentations \cite{laskin2020reinforcement}, such as crop, flip, cutout, and rotation are used to improve the generalization ability when trains the RAMAVT. Two metrics are adopted to measure active visual tracking performance, that is, episode length and episode reward. In this work, we utilize an agent that takes action in random as baseline and the DRLAVT algorithm as comparison. 

All the training curves of two active visual trackers with different inputs are depicted in Fig. \ref{fig3}. We find that the learning progresses of DRLAVTs are much faster than RAMAVTs whatever the input format is, because of its simple Q-network architecture and fully observable state. In addition, the depth information contained in inputs are significant for both active visual trackers to achieve higher episode length. It is worth noting that the depth map and color image adopted in this work are normalized to $[0, 1]$.

The whole evaluation results are summarized in Table \ref{table1}. It clearly shows that the RAMAVT taken depth map as input achieves the highest average episode length about 959.1 score with the best real-time performance, which means that our method can quickly track the target for a longer time. Meanwhile, the higher minimum episode length also proves the stability of RAMAVT. Although, its tracking accuracy (i.e. average episode reward) is slightly lower than DRLAVT. We think it results from the inaccurate target's states, such as target position and velocity, estimated by RAMAVT based on recurrent neural network. This problem gets worse when the agent is only allowed to use color images, as shown in the final row of Table \ref{table1}.

We visualize the tracking results of RAMAVT in Fig \ref{fig5}. It can be seen from Fig \ref{fig5_a} that our method can precisely track the target in the whole episodes. In particular, the tracking errors in X, Y, and Z axes rapidly shrink to 0 and oscillate in a small range, as shown in Fig \ref{fig5_b}. The noncooperative target is also steadily kept in the center of FOV after 200 frames (see in Fig \ref{fig5_c}), even it moves fast with high-speed rotation.

Furthermore, we evaluate the RAMAVT tracker under three types of perturbations, involving actuator noise, time delay, and image blur, to show the robustness of our method. The experiment results are listed in Table \ref{table2}. It is obvious that all the three perturbations have influences on the active visual tracking performance in terms of tracking period and accuracy, especially for the time delay which decreases the average episode length (AEL) about $37.4\%$. We think it is because of the inconsistency of target velocity between the training and evaluation, introduced by the random time delay. In the training stage, the target velocity is set as a random constant in one episode, which has naturally been learned by our RAMAVT model. When the 3 types of perturbations work simultaneously, our method is much robuster than the DRLAVT which has 124.5 scores less under AEL metric.


  
   


\begin{figure}
   \centering
   \includegraphics[width=0.4\textwidth]{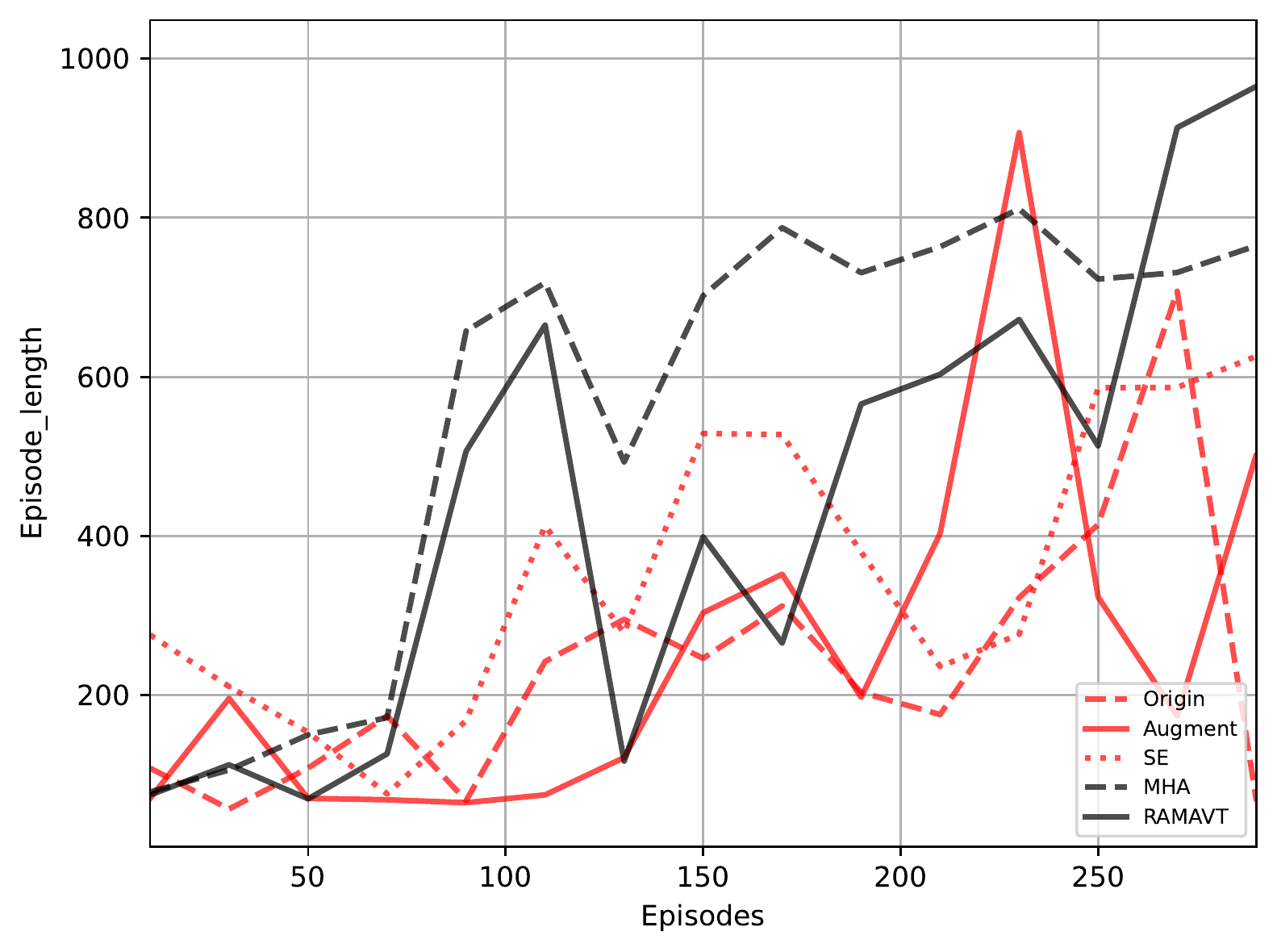}
   \caption{The training curves of ablation models}
   \label{fig7}
\end{figure}

\begin{table}
   \centering
   \caption{Ablation Study on RAMAVT with RGBD image}
   \label{table3}
   \begin{tabular}{cccccccc}
      \toprule
      Name  & AEL & AER  & Speed \\
      \midrule
      Origin & 368.5 & -1876.3  & 214.3 \\

      Augment & 419.3 & -1283.6 & 215.7 \\
  
      SE & 625.8 & -956.0 &  211.3 \\
   
      MHA & 731.1 & -845.7 & 206.6 \\
      \midrule
      RAMAVT & \textcolor{green}{952.4} & \textcolor{green}{-398.5} & 202.7  \\
      \bottomrule
   \end{tabular}
\end{table}

\begin{figure*}
   \centering
   \subfloat[Input image]{\includegraphics[width=0.18 \textwidth]{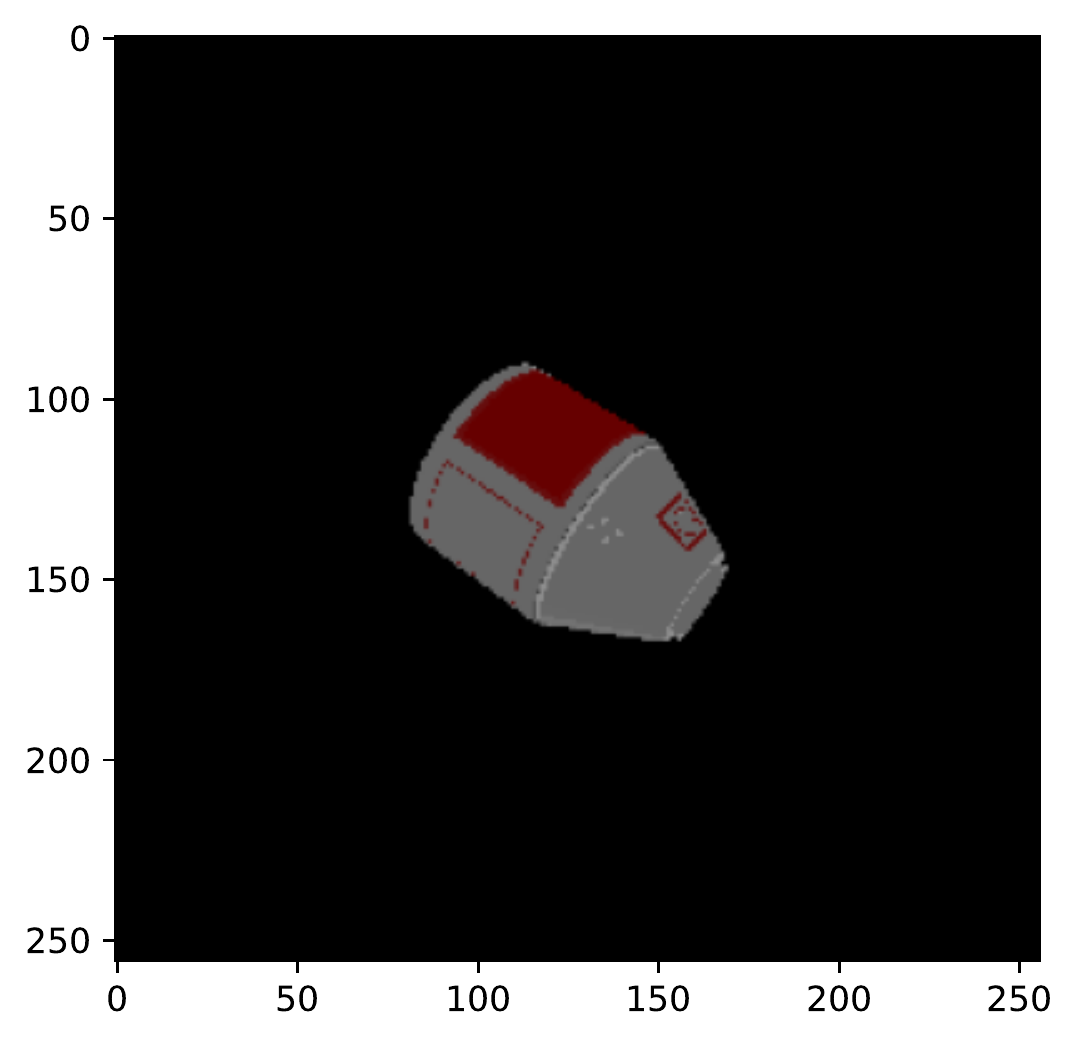}\label{fig6_a}} 
   \subfloat[1st Conv block]{\includegraphics[width=0.18 \textwidth]{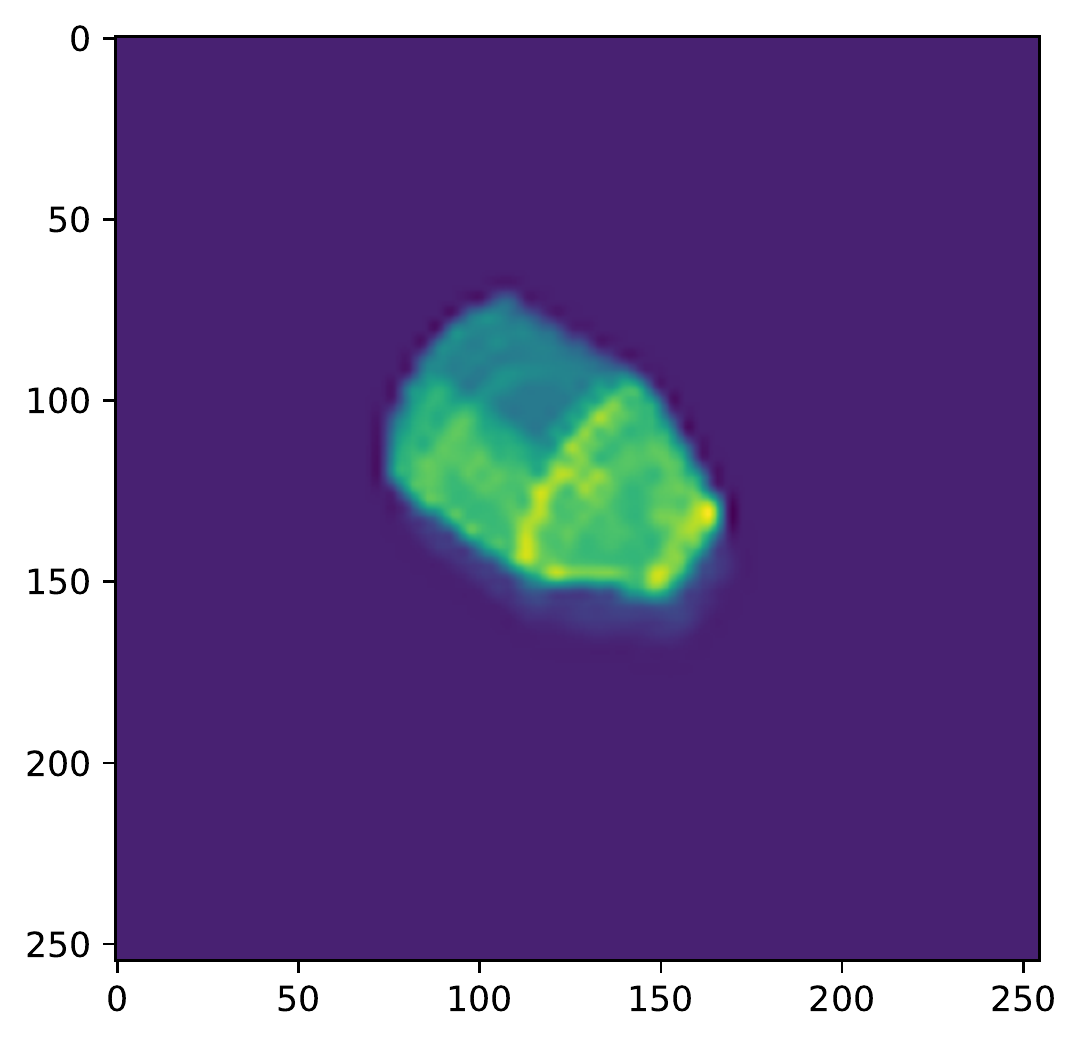} \label{fig6_b}} 
   \subfloat[2nd Conv block]{\includegraphics[width=0.18 \textwidth]{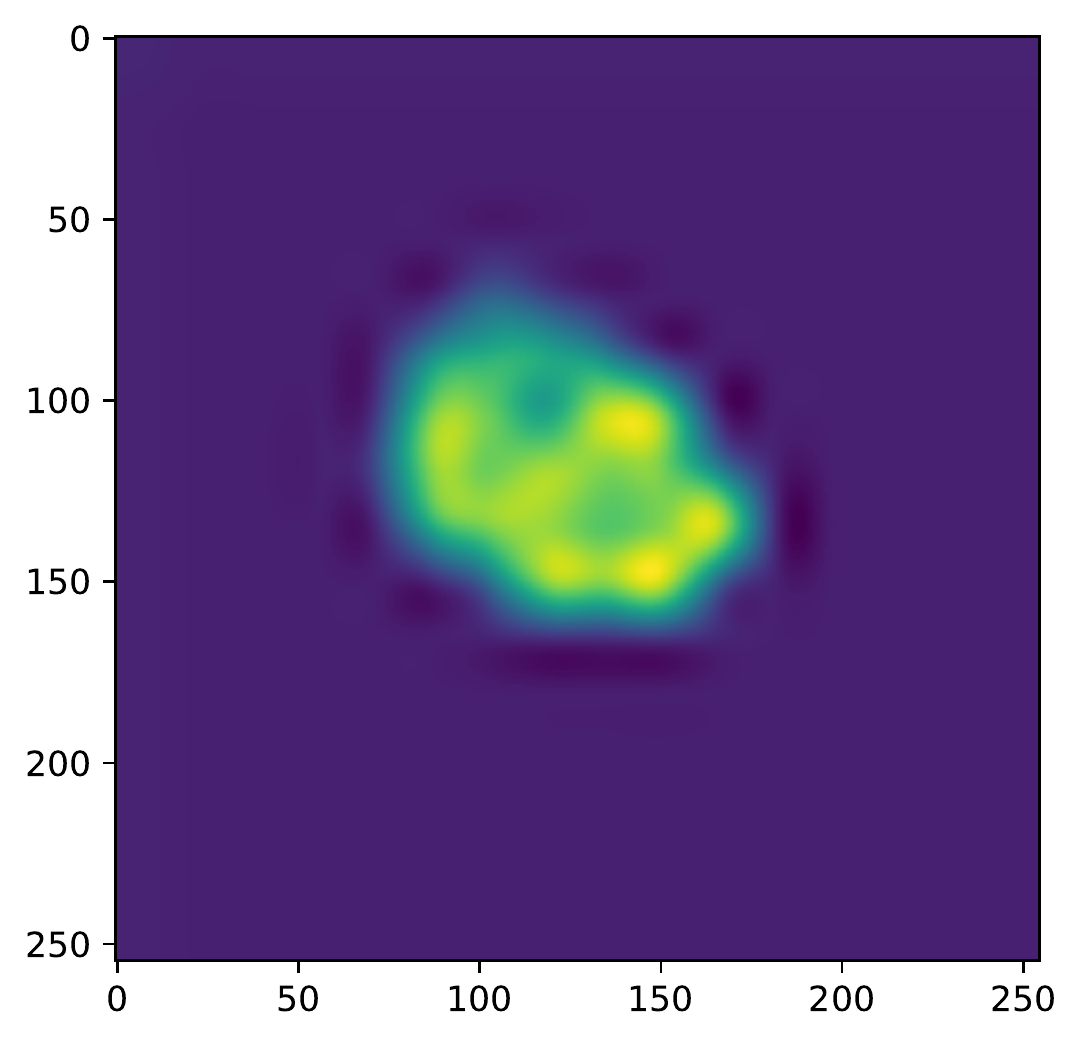} \label{fig6_c}} 
   \subfloat[3rd Conv block]{\includegraphics[width=0.18 \textwidth]{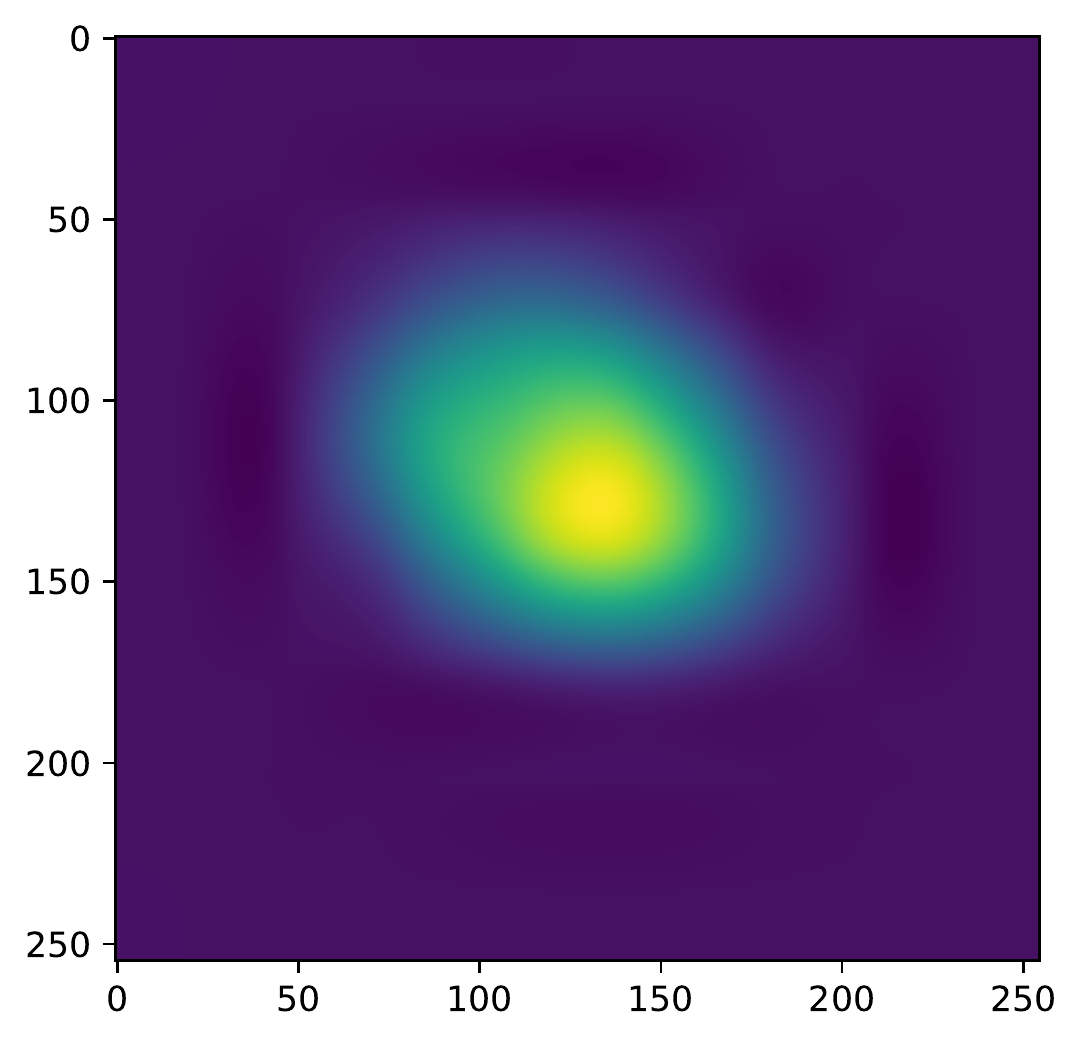} \label{fig6_d}} 
   \subfloat[4th Conv block]{\includegraphics[width=0.18 \textwidth]{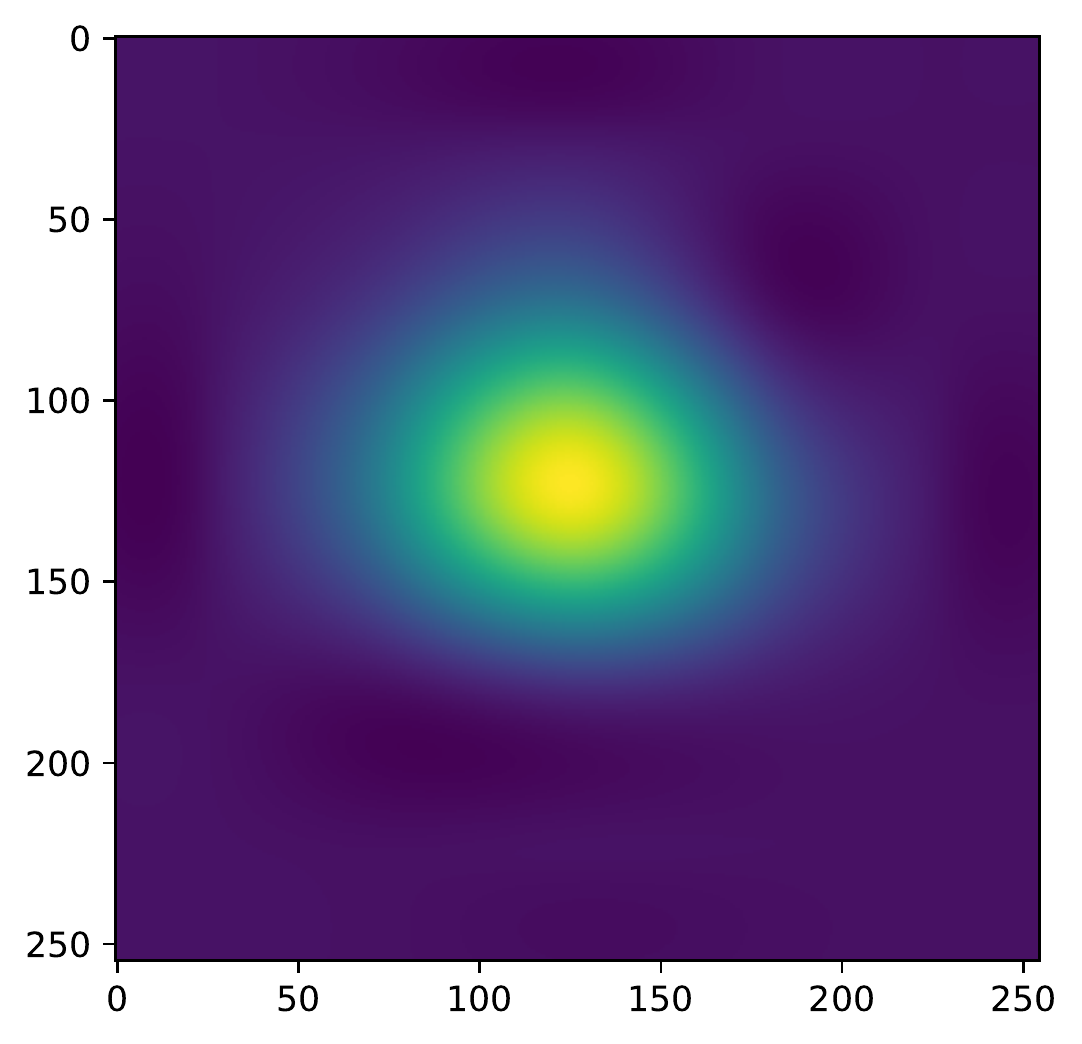} \label{fig6_e}} 
   \\
   \subfloat[Input image]{\includegraphics[width=0.18 \textwidth]{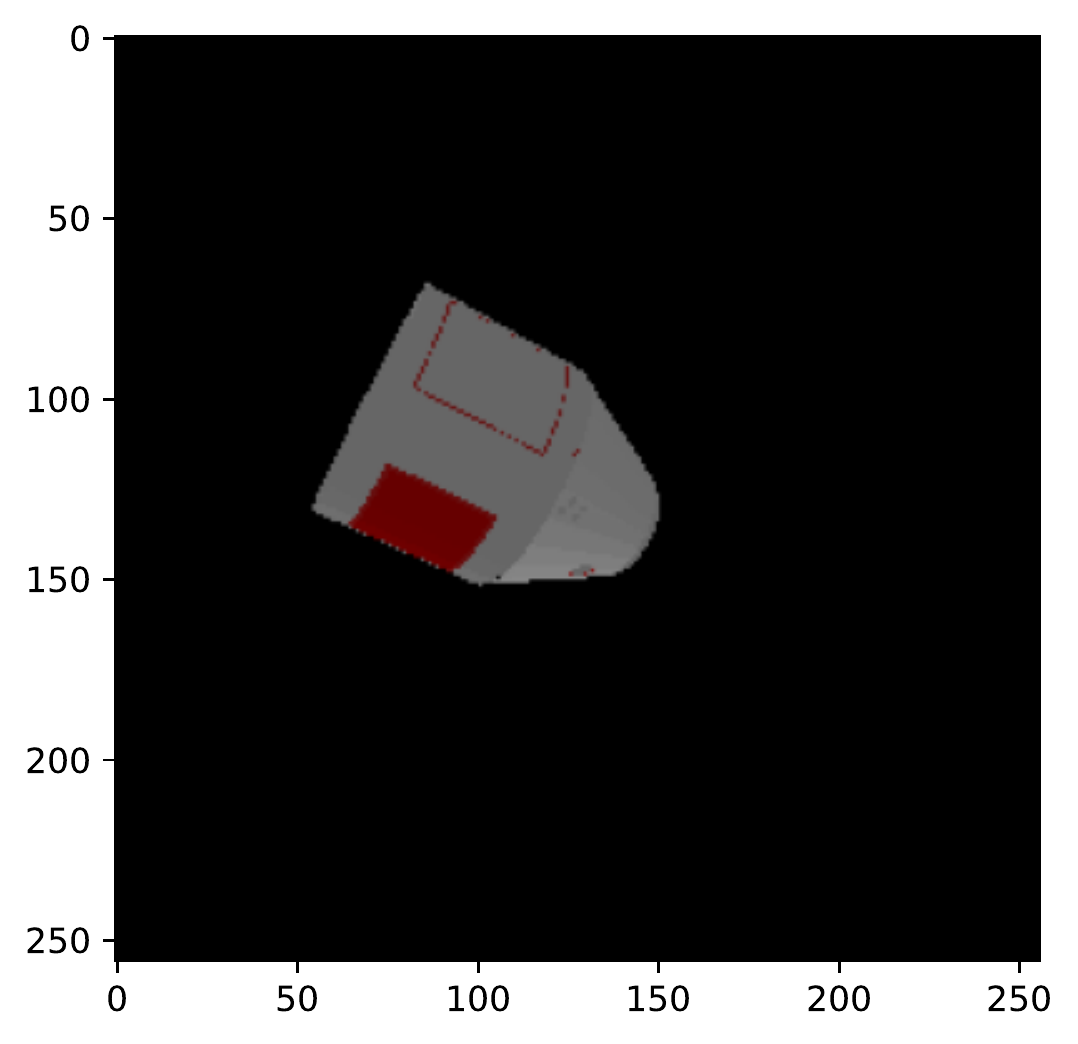} \label{fig6_f}} 
   \subfloat[1st Conv block]{\includegraphics[width=0.18 \textwidth]{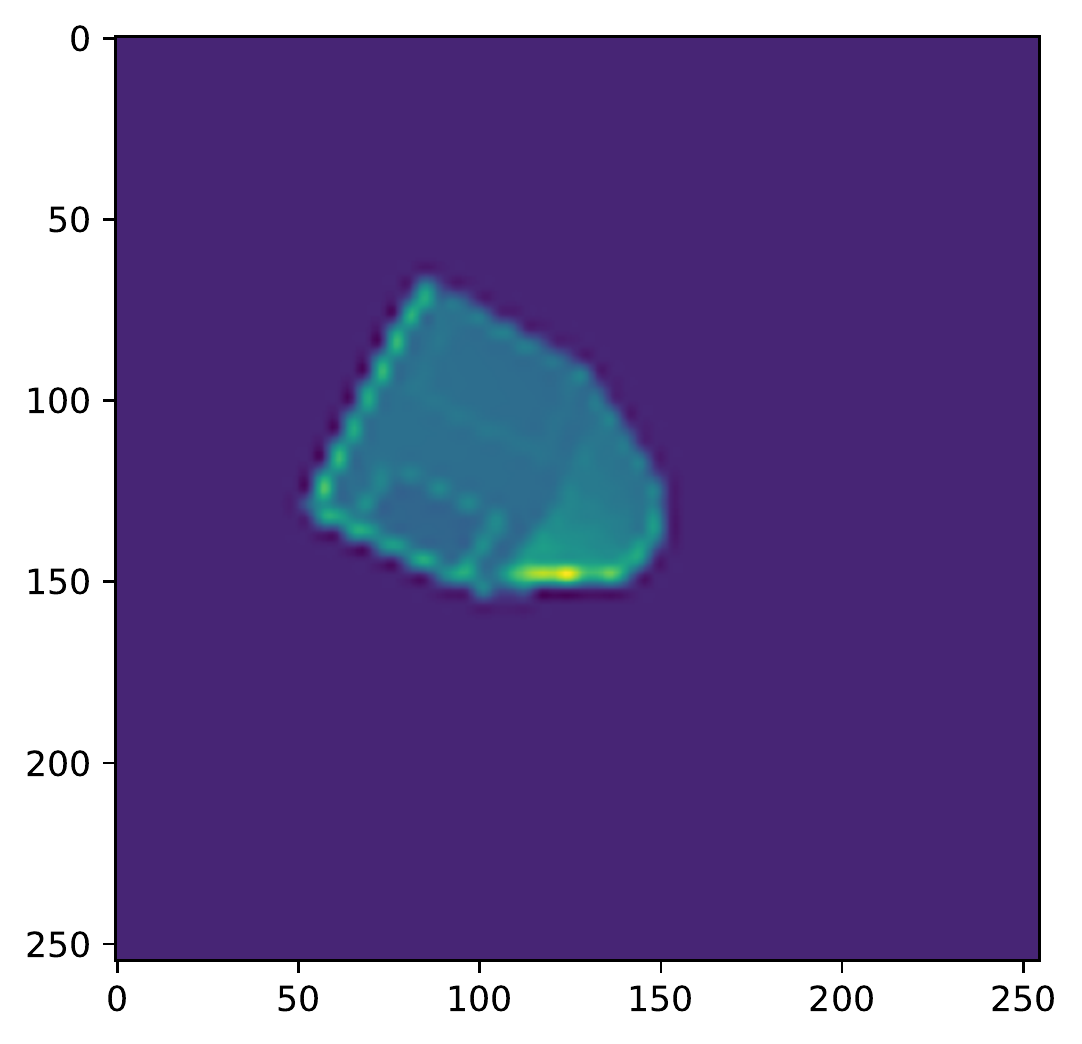} \label{fig6_g}}
   \subfloat[2nd Conv block]{\includegraphics[width=0.18 \textwidth]{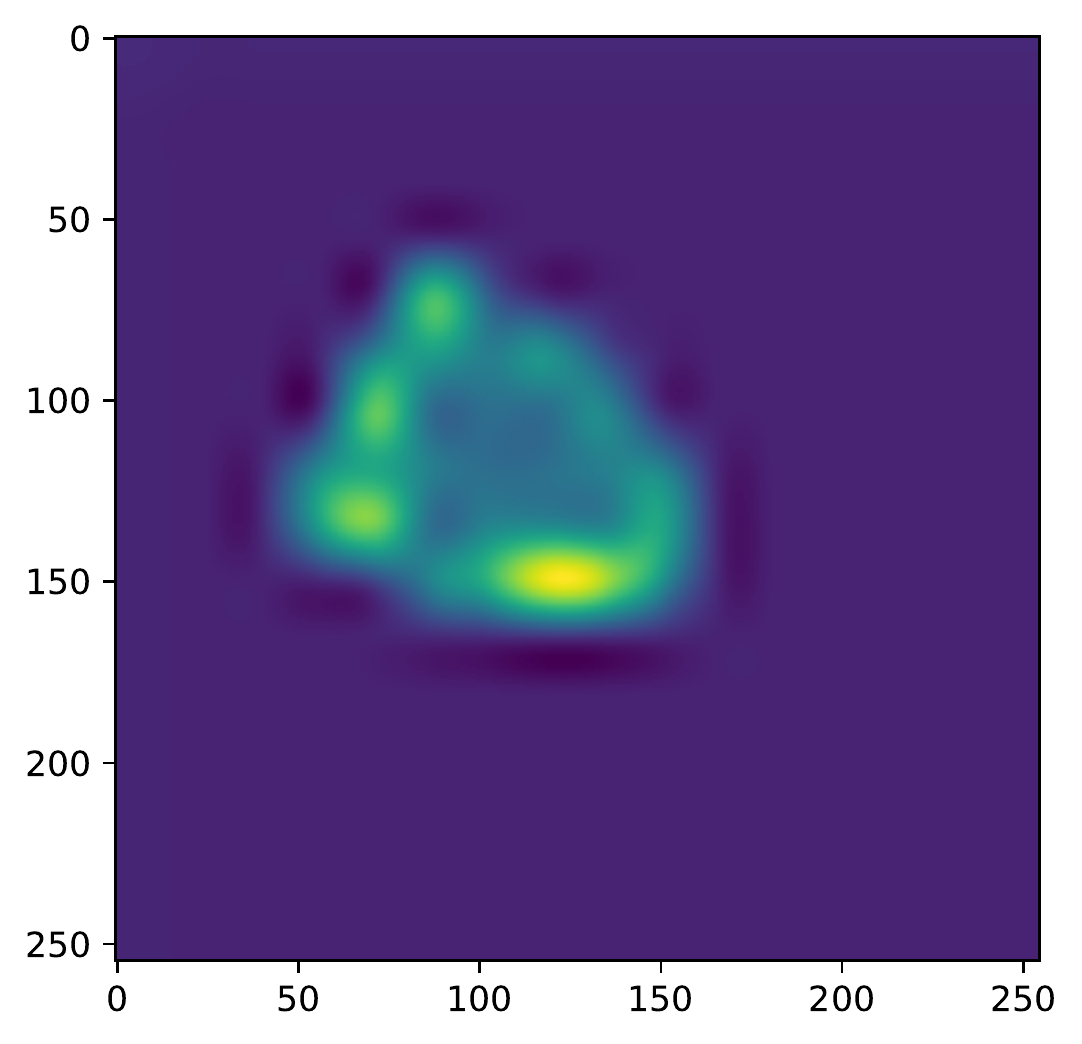} \label{fig6_h}}
   \subfloat[4th Conv block]{\includegraphics[width=0.18 \textwidth]{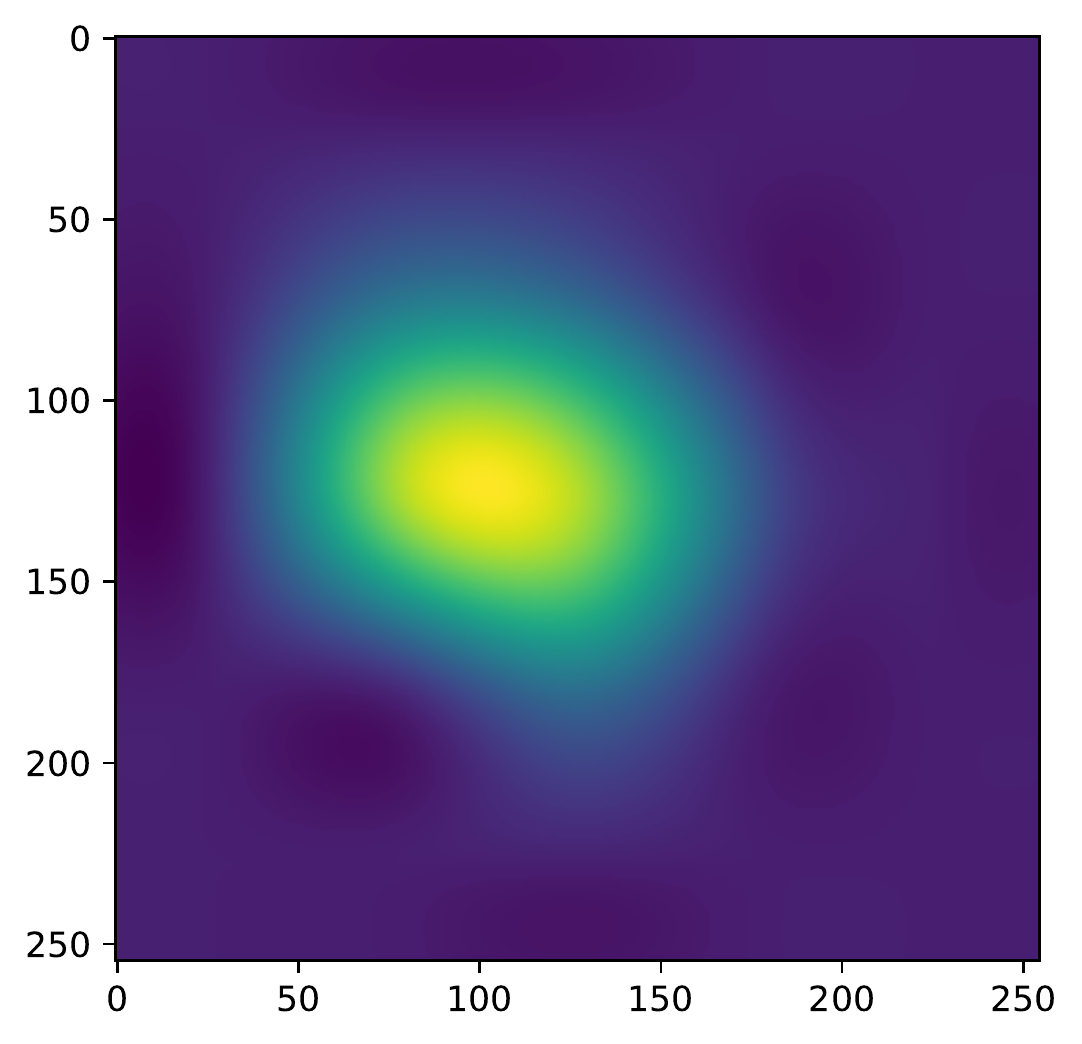} \label{fig6_i}}
   \subfloat[MHA module]{\includegraphics[width=0.18 \textwidth]{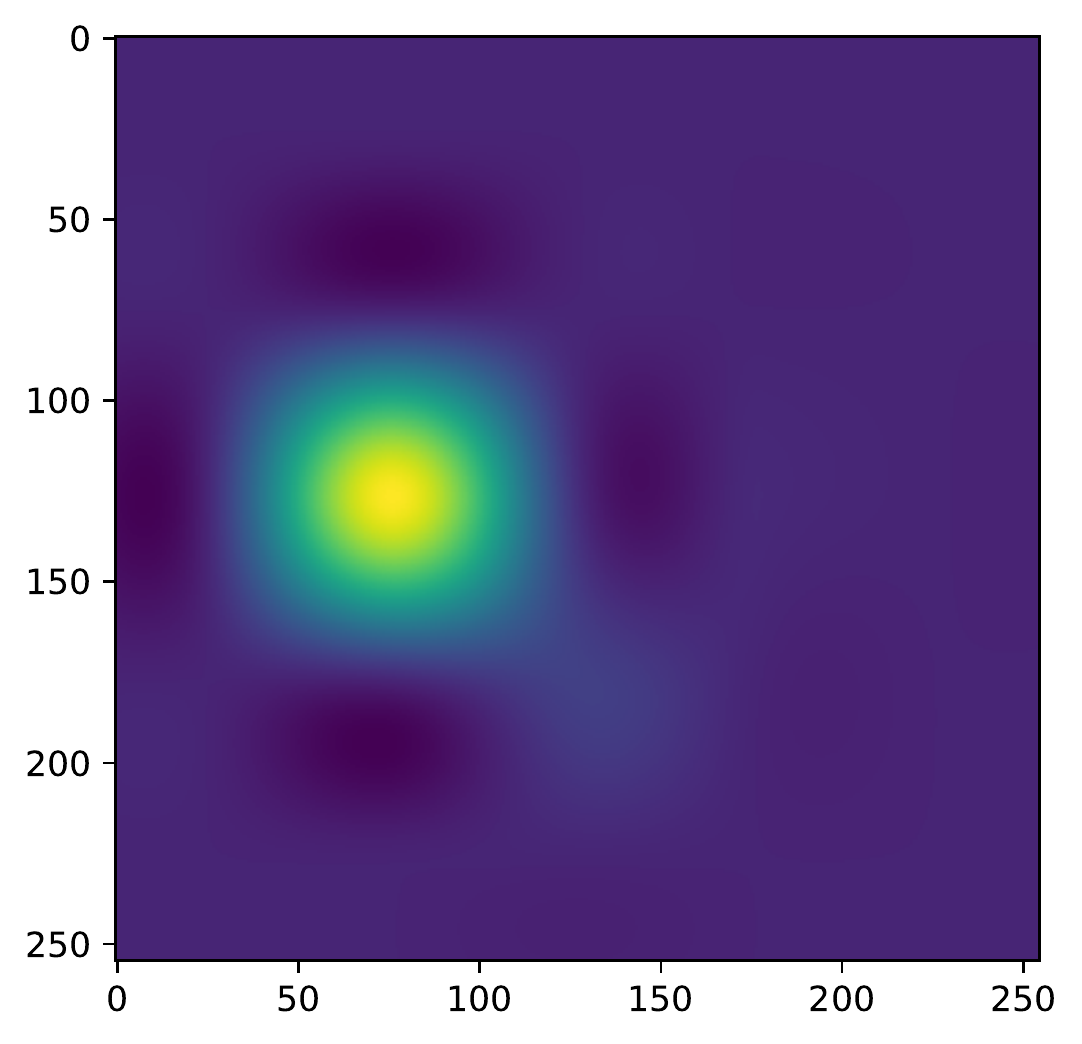} \label{fig6_j}}

   \caption{The interpretability research of active visual trackers. The feature tensors of DRLAVT and RAMAVT are respectively visualized in the first and second rows. It is worthwhile noting that only the first frame of stacked input is depicted at \ref{fig6_a}.}
   \label{fig6}
\end{figure*}

\subsection{Ablation Study}
To show the effectiveness of RAMAVT model, we respectively add data augmentations, SE layer, and MHA module to the original DRQN architecture \cite{hausknecht2015deep}. All the ablation models are trained from scratch with the same training configurations, of which training curves are illustrated in Fig \ref{fig7}. It can be clearly seen that the MHA module not only accelerates the learning progress of agent, but also significantly improves the episode length. This advancement attributes to the attention mainly focusing on the target, which makes the agent more sensitive and accurate to the movement of target. 

The final evaluation results of ablation models are summarized in Table \ref{table3}. It proves that the MHA module achieves the highest AEL and AER scores compared to the others, which only decreases $3.6\%$ running speed. The SE layer also increases the AEL measurement up to 1.7 times with almost no real-time performance loss. In addition, the data augmentation algorithms adopted in this paper including crop, cutout, flip, and rotation work unsatisfactorily, although it does not induce any computational burden during evaluation stage. 

In a word, the RAMAVT model proposed in this work achieves excellent active visual tracking performance in less computational cost, mainly benefitted from spatial-wise and channel-wise attention mechanism induced by the MHA module and SE layers. 

\subsection{Interpretability Research}
We utilize the neural network interpretability method \cite{zagoruyko2017paying} that summarizes the squares of activation values along channel-wise axis and follow with 2-D Softmax operation to explore the inner working mechanism of active visual trackers. The features extracted by each layer of neural network are separately visualized in Fig \ref{fig6}.

The Fig. \ref{fig6_a}-\ref{fig6_e} illustrates different levels of features extracted by DRLAVT of which architecture only contains 4 convolutional blocks. Each convolutional block involves a convolutional layer, a batch-normalization layer, and ReLU activation function. It is worthwhile noting that DRLAVT stacks 4 consecutive frames in channel-wise as one input, however, only the first frame is depicted at Fig. \ref{fig6_a}. We clearly see that the first convolutional block extracts all the edges of target and generates higher feature value to the white body of noncooperative target. The second convolutional block further enhances the reactions to parts of edges. In the subsequent convolutional blocks, more high-level features without specific implications are extracted. The final output of ConvNet backbone looks like a point light source that follows normal Gaussian distribution.

In comparison, the visualization of RAMAVT is not the same, because of two significant differences between the backbone of RAMAVT and DRLAVT: (1) SE layer is added into the first three convolutional blocks, (2) the ConvNet backbone follows with a MHA module. Therefore, the first convolutional block no longer focus on object color (see Fig. \ref{fig6_b} and \ref{fig6_g}). The second convolutional block is more interested in the contour of noncooperative target. Furthermore, the distribution of final output turns more compact caused by the MHA module, which helps to estimate more accurate action value.

\section{Conclusion\label{section5}}
In this paper, we formulate the active visual tracking task of space noncooperative object as POMDP problem and propose a novel active tracker based on deep recurrent reinforcement learning, RAMAVT of which architecture creatively adopts Squeeze-and-Excitation layer and Multi-Head Attention module. It can guide the chasing spacecraft to approach arbitrary space noncooperative target with optimal and high-speed velocity control commands. The advancement of RAMAVT has been proved by sufficient experiments, compared to the state-of-the-art method DRLAVT. To show the effectiveness of our method, we implement convincing ablation study on RAMAVT architecture. In addition, we further take an interpretability research on two active visual trackers to explore their inner working mechanism.






\bibliographystyle{IEEEtran}
\bibliography{IEEEabrv,Reinforcement_Learning,Object_Tracking,My_Work,Visuomotor_control,Active_Visual_Tracking}


\end{document}